\documentclass[sn-mathphys]{sn-jnl} 

\usepackage{amsmath,amssymb,amsfonts}
\usepackage{algorithm}
\usepackage{algorithmicx}
\usepackage{bm}
\usepackage{subfig}
\usepackage{graphicx}
\usepackage{xcolor}

\jyear{2021}%

\theoremstyle{thmstyleone}%
\theoremstyle{thmstyletwo}%

\theoremstyle{thmstylethree}%

\begin{document}

\title[Automatic Detection of Industry Sectors]{Automatic Detection of Industry Sectors in Legal Articles Using Machine Learning Approaches}


\author[1,2]{\fnm{Hui} \sur{Yang}}\email{hy20248@essex.ac.uk}

\author[1]{\fnm{Stella} \sur{Hadjiantoni}}\email{stella.hadjiantoni@essex.ac.uk}

\author[3]{\fnm{Yunfei} \sur{Long}}\email{yl20051@essex.ac.uk}

\author*[2]{\fnm{R\={u}ta} \sur{Petraityt\.{e} }}\email{ruta.petra@mondaq.com}

\author*[1,4]{\fnm{Berthold} \sur{Lausen}}\email{berthold.lausen@fau.de}

\affil[1]{\orgdiv{Department of Mathematical Sciences}, \orgname{University of Essex}, \orgaddress{\street{Wivenhoe Park}, \city{Colchester}, \postcode{CO43SQ},  \country{UK}}}

\affil[2]{ \orgname{Mondaq Ltd},  \orgaddress{\city{Bristol}, \country{UK}}}

\affil[3]{\orgdiv{School of Computer Science and Electronic Engineering}, \orgname{University of Essex}, \orgaddress{\street{Wivenhoe Park}, \city{Colchester}, \postcode{CO43SQ},  \country{UK}}}

\affil[4]{\orgdiv{Institute of Medical Informatics, Biometry and Epidemiology, School of Medicine}, \orgname{Friedrich-Alexander University Erlangen-Nuremberg},
    \orgaddress{\street{Waldstr. 6}, \city{Erlangen}, \postcode{91054},  \country{Germany}}}


\abstract{The ability to automatically identify industry sector coverage in articles on legal developments, or any kind of news articles for that matter, can bring plentiful of benefits both to the readers and the content creators themselves. By having articles tagged based on industry coverage, readers from all around the world would be able to get to legal news that are specific to their region and professional industry. Simultaneously, writers would benefit from understanding which industries potentially lack coverage or which industries readers are currently mostly interested in and thus, they would focus their writing efforts towards more inclusive and relevant legal news coverage. In this paper, a Machine Learning-powered industry analysis approach which combined Natural Language Processing (NLP) with Statistical and Machine Learning (ML) techniques was investigated. A dataset consisting of over 1,700 annotated legal articles was created for the identification of six industry sectors. Text and legal based features were extracted from the text. Both traditional ML methods (e.g. gradient boosting machine algorithms, and decision-tree based algorithms) and deep neural network (e.g. transformer models) were applied for performance comparison of predictive models. 
The system achieved promising results with  area under the receiver operating characteristic curve scores above 0.90 and F-scores above 0.81 with respect to the six industry sectors. The experimental results show that the suggested automated industry analysis which employs ML techniques allows the processing of large collections of text data in an easy, efficient, and scalable way. Traditional ML methods perform better than deep neural networks when only a small and domain-specific training data is available for the study.}




\maketitle

\section{Introduction}\label{sec:intro}

Automatic text classification aims at organizing diverse and unstructured textual data by categorizing them, based on their content, in a set of pre-defined classes. With the vast amounts of textual documents in digital form nowadays, which constantly increase, text classification has become an essential part of content processing. Unstructured textual data may take the form of emails, news articles, business documents or customer reviews.  Automatic text classification is essential for information extraction, summarization and text retrieval, and finds applications in the digital document management, spam filtering, sentiment and opinion mining to name but a few \cite{dalal:2011}.  

In the legal domain, massive amounts of unstructured text data are generated or collected from documented legal cases, 
communication with clients, or published legal articles \cite{sulea:2017,wei:2018}. Similarly to social media which are nowadays used by businesses for online marketing purposes to connect with customers \cite{hoang:22}, when exploring vast amounts of diverse and unstructured textual data, effective access to the required information is likely appreciated not only by professionals in the legal domain but also by regular legal news followers.
On platforms like Mondaq\footnote{https://www.mondaq.com}, that host hundreds of thousands of news articles related to a variety of legal topics, 
it is essential that articles are appropriately tagged and categorized in order to help readers find the information they are looking for. 
Specifically, as the volume and diversity of content grows, manually labelling articles every time a new label is introduced to the overall content 
taxonomy (like the industry label) can be quite time consuming and expensive. Therefore, it is necessary 
to explore predictive approaches, with the help of automatic text processing, 
since it often requires a relatively small sample of labelled examples to train on and which can then be used to label the rest of the data automatically.

Typically, researchers categorize text processing into traditional models and deep learning models. 
However, since the 2010s, text classification has gradually changed from traditional models to deep 
learning models, especially with the popularity of transformer models after 2018. Compared with the 
methods based on traditional machine learning, deep learning methods avoid designing rules and features
 by humans and automatically provide semantically meaningful representations for text mining. Most of the 
 current text classification research works are based on Deep Neural Networks (DNNs).

However, DNNs are not magic bullets for every task. Firstly, DNNs always come with high computational 
complexity and require more data, which might cause  problems in the predictive task, as for example the 
data availability may be limited. 
Secondly, it is always tricky to incorporate additional knowledge into DNNs. Although many researchers are working on this, 
see for example \cite{du:21}, developing that approach might not be affordable for small to medium sized companies or not 
applicable in a specific domain like legal services. In addition, DNNs are generally regarded as a black-box approach, 
while the explainable DNNs are still in the infant stage. Hence it is still necessary to investigate how traditional machine 
learning can be used on industry-specific text classification without compromising model performance.

In this paper, the problem of automatic text classification for the categorization of legal articles according to the 
industry sector topic, herein referred to as the \emph{industry sector detection problem}, is examined using traditional 
machine learning (ML) and deep learning (DL) models. The aim is to detect multiple industry sectors hidden in massive 
texts by mining underlying semantics within these texts.  
Specifically, the first objective of this study is as follows: given a document, content annotators would have to make 
complex decisions, including labelling the articles based on the industry sector they refer to.  
Given the available data from previous manual industry sector detection analysis, is it possible to train text 
classification systems which can predict the decisions that would be made by the annotators? Such a system 
could act as a decision support system or a sanity check for content annotators. The second objective is to build 
a simple machine learning model with moderate feature engineering for the industry sector prediction task. The 
proposed model, based on traditional ML models with mild feature engineering, hopes to achieve lower running 
time and better explainability while it does not compromise predictive performance.

To achieve the above two objectives and fulfill the current research gap, a small domain-specific 
dataset, related to different areas of law with more than 1,700 legal articles, was firstly curated by manually 
labelling the articles with six industry sectors, so that it can be used to build the traditional ML classifier-based 
predictive models. 
Then, some useful legal information (e.g. legal topics) was introduced to this study to enrich further the generation of the 
feature set for the ML approaches, thus helping explore the implicit association between legal information and industry 
sectors. Our experiments compare different ML algorithms, including a wide range of traditional ML and DNNs. Our 
experimental results demonstrate that with feature engineering, traditional ML models like RGF outperform DNNs 
which run much slower than RGFs even using GPU.
Meanwhile, the proposed model has enhanced explainability and better fits the legal sector's industry requirements.

The remaining of the paper is organised as follows. Section~\ref{relatedworks} provides a short literature review on 
the use of machine learning and deep learning models in text classification problems. Details about the dataset used for 
industry sector detection problem are provided in Section~\ref{material}. Section~\ref{method} presents the 
proposed system framework which has been developed to tackle the complexity of the industry sector detection problem. 
The experimental results are presented in Section~\ref{results}. Section~\ref{discussion}, discusses some implications 
arising from the feature selection stage, provides an overview of the employed ML and DL algorithms' performance, and also 
discusses error analysis. Finally, Section~\ref{conclusion} concludes and outlines some topics for future work.

\section{Related works}
\label{relatedworks}

Legal service related industry detection is closely related to text classification, which is the most fundamental and essential 
task in natural language processing. Although industry detection is rarely mentioned in the research, other text classification 
applications, such as sentiment analysis \cite{long-etal-2017-cognition,shen2020dual}, topic labeling \cite{bechara2021transfer}, 
deception detection \cite{long2017fake,shu2017fake} and dialog classification \cite{lee2016sequential} are very popular.

Until the 2010s, text classification mostly includes tree-based or statistic-based algorithms. Statistic based methods, like Naïve Bayes 
(NB) \cite{pang-etal-2002-thumbs}, K-Nearest Neighbor (KNN) \cite{han2001text}, Random Forests (RF) \cite{breiman:2001} and 
Support Vector Machine (SVM) \cite{colas2006comparison}, aim to obtaining statistical significant features based on probabilities.

Since 2010, impacted by the emergence of word embedding \cite{levy2015improving}, numerous deep learning models have been 
proposed for text classification. The recursive neural network \cite{chen2015sentence} is the first deep learning approach used for 
text classification tasks, which improves performance compared with traditional models. Then, Convolutional Neural Network (CNNs), 
Recurrent Neural Networks (RNNs)(and its variants like LSTM/GRU), and Memory Networks have been used in text classification. 
The contextualized word embedding and transformer, like  Embeddings from Language Models (ELMo) \cite{ilic2018deep} and 
BERT \cite{devlin2018bert}, is a significant turning point in the development of text classification and other NLP applications. 
These are often called pre-training models and they can generate contextualized word vectors.

Compared to traditional machine learning models, deep learning models can learn feature representations directly from the 
input without intensive manual feature engineering and knowledge intervention. However, deep learning technology usually 
requires enormous data (often hundreds of thousands) to achieve high performance. Moreover, in terms of the ability to be 
interpolated, although attention-based models can bring some interpretability among words for DNNs, it is not enough
 compared with traditional models to explain why and how it works well. In addition, DNNs often require a cluster of GPUs 
 and training for hours and days, which is not always affordable for smaller and medium sized companies, and not relatively 
 environmentally friendly in general. Hence, it is still necessary to investigate traditional machine learning methods in many 
 text classification tasks.

\section{Data Collection}
\label{material}

The unstructured, textual data consist of articles on legal topics 
which contain information on a wide variety of industry sectors. Such information provides detailed insights into how 
specific industries or markets are performing and/or are likely to perform in one particular region or period of time in the future.

Initially, a set of 28 industry sectors were identified from the articles during the data curating process. After the annotation stage, six industry sectors were selected based on two reasons: Firstly they are popular industries and frequently appear in relevant legal documents. Secondly, in the curated training data stage, these industry sectors were annotated in more articles than other industry sectors. The selected industry sectors are as follows:

\begin{itemize}
\item Financial Services: \textit{Banking \& Credit,} \textit{Securities \& Investment}
\item Health: \textit{Healthcare}, \textit{Pharmaceuticals \& BioTech}
\item Technology
\item Property
\item Energy: \textit{Oil \& Gas}, \textit{Utility}
\item Insurance
\end{itemize}
It is noted that several closely related industries are merged into a more general industry. For example, the \textit{Financial Services} industry consists of two subcategories, namely \textit{Banking \& Credit}, and \textit{Securities \& Investment}.

\begin{figure}
\includegraphics[scale=.8]{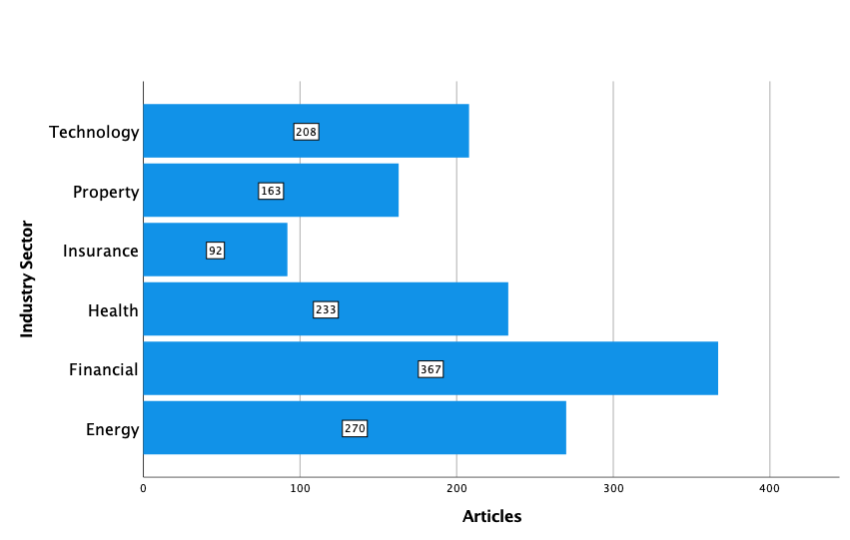}
\caption{The distribution of the 124 out of 1042 articles with co-occurred industry sectors.}
\label{fig:industry_article}       
\end{figure}

A total of 1,730 full-text articles were curated by the annotators, and 1,355 (78.3\%) articles were annotated with at least one industry tag 
out of the original 28 industry sectors. The remaining 21.7\% of annotated articles were not assigned to belong to an industry sector, which 
are then removed from the dataset. In the remaining data with 1,355 articles, 1,042 articles were labelled with at least one of the six industry 
sectors studied here. Figure~\ref{fig:industry_article} displays the number of the articles which were tagged to belong to each of the target 
industry sectors. These tags will be treated as the positive instances for the building of the predictive models, that is, they are considered 
the labels of the collected articles.

Moreover, an article might be relevant to more than one selected industry. In this dataset, 124 articles were assigned with two or more 
target industry sectors. Table~\ref{tab:article_coocurrence} shows the co-occurrence of the six industry sectors in the 1,042 articles used. 

\begin{table}[t]
\centering
\begin{tabular}{|lllllll|}
\hline
           & Financial & Health & Technology & Property & Energy & Insurance \\\hline
Financial  & -         & 15     & 28         & 11       & 12     & 16        \\
Health     & -         & -      & 34         & 4        & 4      & 10        \\
Technology & -         & -      & -          & 2        & 9      & 4         \\
Property   & -         & -      & -          & -        & 3      & 8         \\
Energy     & -         & -      & -          & -        & -      & 3         \\
Insurance  & -         & -      & -          & -        & -      & -         \\\hline
\end{tabular}
\caption{The distribution of the articles with co-occurred industry sectors.}
\label{tab:article_coocurrence}       
\end{table}

\section{Methodology}
\label{method}

In this paper, six main industry sectors were chosen for the industry sector detection classification problem. The industry sector detection 
task is considered as a binary classification problem where for each industry sector, a separate hybrid predictive model is built by text 
classification algorithms with the aim of determining whether an article is associated with a particular industry or not. 

\begin{figure}[h]
\includegraphics[scale=.75]{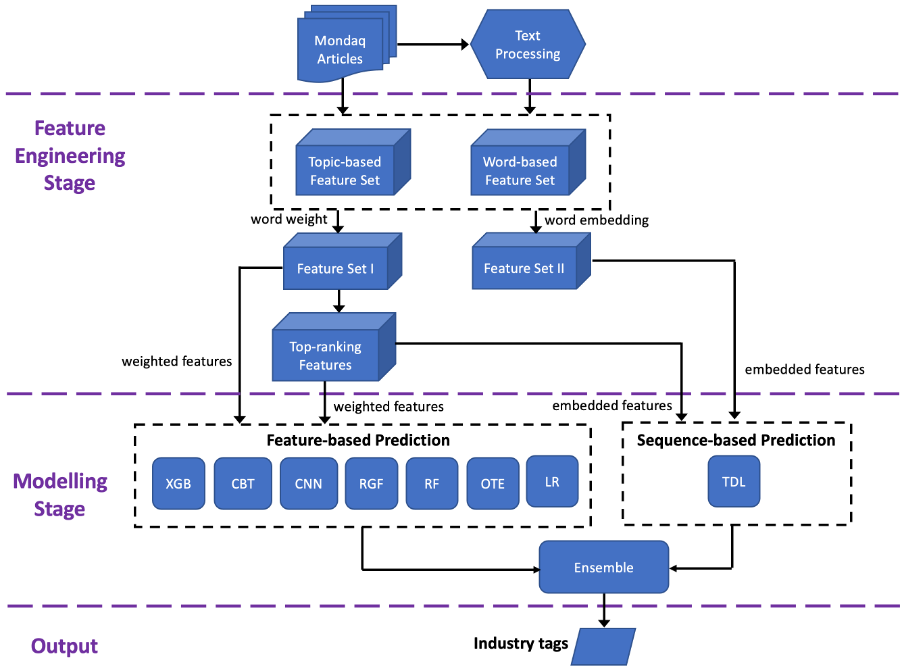}
\caption{System framework for the ML-based industry detection. 
}
\label{fig:system_framework}      
\end{figure}

The system framework (Figure~\ref{fig:system_framework}) for the industry detection task comprises two main stages, the feature 
engineering stage and the modeling stage. In the feature engineering stage, given a set of full-text articles, two types of features are 
firstly obtained from the articles: one is word-token based features extracted from the full text via a series of text processing steps, 
and the other is the article's legal topic tags which are directly provided by the article editors. Then, two feature sets are separately 
generated: one is Feature Set I with one-dimension weighted features, and the other is Feature Set II with high-dimension features 
generated by word embedding techniques. The two feature sets which are created to target different prediction approaches. 
Furthermore, top-ranking features are selected from the Feature Set I to improve prediction performance. Following the feature engineering 
and selection stages, a number of classifiers are then applied for performance comparison.

\subsection{Text pre-processing}\label{subsec:text_process}

For the full-text articles, a number of NLP steps are employed in order to obtain word-based features used for the ML-based 
model building. Firstly, unimportant sections in the articles such as `FootNote' and `Disclaimer', which are located at the end 
of the articles, are removed from the text. Then, text cleaning work is conducted by deleting the HTML tags (e.g. $<$html$>$) 
and special characters (e.g. \&amp) from the text. Punctuation and stop words (e.g. a, the) are also discarded from the text.

\subsection{Proposed features}
\label{subsec:feature_set}

The feature set used for the construction of the ML-based predictive models consists of two main parts: one is word tokens 
extracted from the text (see Section~\ref{subsec:text_process}) and the other is topic-based features with distinct legal topic 
tags. Specifically, each legal article was associated with a number of legal topic tags which were provided by Mondaq editors 
when they manually examined the article content. A set of 206 topic tags were created to cover the wide variety of legal topics 
(e.g \textit{Aviation Finance}, \textit{Insolvency} and \textit{Litigation}) regarding legal, accounting and consultancy domain areas. 
A set of parent-child relations were created to demonstrate the potential associations between parts of the topic tags. For example, 
the topic `\textit{Accounting and Audit}' has several children topics, that is '\textit{Audit}', `\textit{Taxation}', and `\textit{Forensic Accounting}'. 
In this study, legal topic tags are used as another set of useful features for industry detection.

Two feature sets were separately created for the use of different prediction approaches: 

\textbf{Feature Set I.} In this feature set, each feature is represented with a \emph{one-dimension} vector. Legal topic features are 
binary variables to indicate the association with the target article. For word-token features, word stemming is firstly applied to obtain 
the basic form of the word to help reduce the size of the word features. A total of 40,758 word features are extracted from all the 
annotated articles. Then, for each word token feature, the feature weight value is calculated by either one-hot encoding or the term 
frequency–inverse document frequency measure (\(tf \times idf\)) \cite{Wu:2008}.

\textbf{Feature Set II.} In this feature set, each feature is described as a vector of high dimension that is created by word embedding 
techniques like the Word2Vec method \cite{Mikolov:2013,pennington:2014}. The Word2Vec retains semantic meaning between the 
words to some extent by taking into account word co-occurrence within local context. 

\subsection{Feature selection}
\label{subsec:feature_selection}

It is well known that in statistics and machine learning modelling not all the features are important. Therefore, it is vital to uncover those 
important features with significant effect on label prediction. For Feature set I with one-dimension features, a feature selection stage is 
applied to discover important features in terms of different industry sectors. Since each industry sector has its own distinct characteristics, 
the feature selection step was conducted separately for each industry sector. 
Algorithm~\ref{feature_selection} summarises the feature selection steps in terms of one particular industry sector.

\begin{algorithm}
\caption{Selecting the most important features for each industry sector.}
\label{feature_selection}
\begin{algorithmic}[1]
\State The full feature set is firstly fed into four ML algorithms (including XGB, CBT, RF, and LR).
\State For each ML algorithm, the relevant importance score of each feature is calculated by the built-in feature importance function.  
\State In each ML algorithm, an optimal score threshold is used to uncover the important features that have impact on prediction. 
The optimal score threshold is determined by a set of experiments using different score thresholds. In each experiment, a feature 
importance score threshold is tried and then a set of top-ranking features above this score threshold is created for performance comparison. 
Finally, the optimal threshold is set by the top-ranking feature set with the best performance.
\State For each ML algorithm, the top-ranking features with the importance score above the optimal score threshold are selected from the full feature set. 
\State An integrated feature list is created by combining together the top-ranking feature lists from different ML algorithms. 
\end{algorithmic}
\end{algorithm}

At step~3 of Algorithm~\ref{feature_selection}, the optimal threshold was determined by the experiments based on the prediction 
performance. As we know, different ML algorithms and different feature importance functions might capture different important features 
to each other to some extent. For this reason in step~5 of Algorithm~\ref{feature_selection}, for each industry, an integrated list of 
key features was generated by merging the top-ranking features. That is, the top-ranking features of four different importance measures 
are combined to provide the final set of features that will be fed into the ML models for the building of predictive models with respect to 
one specific industry sector. Specifically, the following four feature importance functions were combined to provide the \textit{Top-ranking 
features Set} shown in Figure~\ref{fig:industry_article}. 
\subsubsection {Mean Decrease Impurity}

Suppose we have a dataset $\mathcal{D}$ and let \( \bm{X} =  ( X_1, \dots, X_p) \in \mathbb{R}^p\) be the set of features 
and \( Y \in \{ 0, 1\} \) be the binary target variable. Assume a forest has resulted from the aggregation of $M$  trees.

In gradient boosting, following \cite{breiman:2001}, importance for a feature is based on a very simple idea: it is the average 
importance over all the trees. Specifically, importance $I_l^2$ for a single feature $X_{l}$ is given by
\begin{equation}
  I_l^2  =  \frac{1}{M}\sum_{l=1}^M I_l^2(T_m),
  \label{eq:MDIXj}
\end{equation}
where the sum of the feature importance of the individual trees $T_m$, $m=1,\dots,M$, is averaged over the total number of
trees $M$ \cite{friedman:2001, hastie:01}. The relative importance of a single tree $T_m$ is calculated by

\begin{align*}
    I^2_l (T_m) = \frac{N_t}{n} ( \text{Impurity}(t) - N_n(t_R) / N_n(t) \text{Impurity}(t_L) \\
                    - N_n(t_L) / N_n(t) \text{Impurity}(t_L)),
\end{align*}
where $N_t$ is the number of samples falling into the subset of current node $t$, $n$ is the sample size, $N_n(t_R)$ and $N_n(t_L)$ 
denote the number of samples falling into the subset of the right child $t_R$ and left child $t_L$ of node $t$, respectively 
and $\text{Impurity}(t)$ is some impurity measure.

As a result, the Mean Decrease Impurity (MDI) calculates feature importance based on the total reduction in weighted node 
impurity contributed by all trees of the ensemble for a given feature. 
 Then, the MDI for feature $X_j$ and a single tree $T_m$ is given by 
 \begin{equation}
    \text{MDI} = \sum\limits_{\substack{t\in T_m,\\ v(t)=X_j}} I^2_l (T_m)
\end{equation}
and for an ensemble of trees it is given by (\ref{eq:MDIXj}).

Herein, in random forests, the impurity measure is the Gini index. The MDI for feature $X_j$ is then given by 
\begin{equation}
    \text{MDI}(X_{j})_g = \frac{1}{M}\sum_{l=1}^M \sum_{\substack{t\in \mathcal{T_l}\\ j^*_{n,t} }} 2p_{n,t} L_{\text{class}, n}( j^*_{n,t}, z^*_{n,t}), 
\label{eq:MDIXjGini}
\end{equation}
where $p_{n,t}$ is the proportion of observations in the node $t$, \( ( j^*_{n,t}, z^*_{n,t}) \) maximises the criterion \( L_{\text{class}, n}\) 
which is evaluated for the number of all possible directions for splitting at each node of the tree \cite{breiman:84}. Specifically, the impurity 
measure in \( L_{\text{class}, n}\) is the Gini index which calculates the probability of misclassification of a legal article to a specific 
industry sector. 

Therefore, in (\ref{eq:MDIXjGini}), the estimated MDI of the feature $X_{j}$ is given as the average weighted decrease of node impurity 
related to splits along the variable $X_{j}$. 

\subsubsection {Feature importance in CatBoost}

In CatBoost, feature importance in the case of non-ranking loss function, is calculated by looking at the average prediction change 
when the value of a feature changes\footnote{https://catboost.ai/en/docs/concepts/fstr\#regular-feature-importance}. The idea is that 
a change in an important feature causes a bigger change in the predicted value compared to a change in a less important feature. 
Consequently, for a feature $F$, feature importance is calculated by
 \begin{align*}
     I_F &= \sum_{\text{trees,leafs}_F} (v_1-avr)^2 c_1+(v_2-avr)^2c_2, \\
     avr &= \frac{v_1 c_1 + v_2 c_2}{c_1+c_2},
 \end{align*}
 where $c_1, \ c_2$ are the total weight of objects, $v_1, \ v_2$ are the the loss function value in the left and right leaves, respectively.

 \subsubsection {Logistic regression}

 Finally, the magnitude of the coefficients of the fitted logistic regression (LR) model, from each industry sector, has been considered 
 as an additional measure of feature importance. Specifically, for a LR model
 \[
logit( \pi_i ) = log \left( \frac{\pi_i}{1-\pi_i} \right) = {\bf x}_i^t \ \boldsymbol{ \beta} + \epsilon_i,
\]
where $\pi_i = P (Y=k \| X = \bf{x})$ with $k=0,1$ for a binary classification problem, 
${\bf x}_i$ is the vector of features 
and  $\boldsymbol{ \beta} $ the vector of coefficients, the fitted model is given by  
\[
logit( \hat{\pi_i} ) = log \left( \frac{\hat{\pi}_i}{1-\hat{\pi}_i} \right) = {\bf x}_i^t \ \boldsymbol{ \hat{\beta}}. 
\]
 The absolute value of each feature's coefficient in $\boldsymbol{ \hat{\beta}}$ is then considered.

As a result, each industry sector has its own top-ranking feature list generated from the above feature selection steps. 
Figure \ref{fig:important_feature} depicts the number of important word and topic features found by the selected 
algorithms with respect to individual industry sectors.

\subsection{Classification algorithms}
\label{subsec:algorithm}

In the modeling stage, we have applied six classification algorithms to investigate their predictability potential on 
industry sector detection. These algorithms were selected for the industry detection classification task because 
they are based on different supervised learning techniques which make use of distinguished mathematical theories. 
In this study, two prediction approaches were employed using separate feature sets (see Section \ref{subsec:feature_set}) 
generated in the feature engineering stage: 

\textbf{Feature-based prediction.} In these feature-based traditional classification, feature variables are treated as independent 
to each other. Feature Set I with one-dimension features was used for this type of industry prediction. A number of traditional 
ML algorithms were applied for the building of predictive models. Specifically, two Gradient Boosting Machine (GBM) \cite{friedman:2001}  
algorithms, \textit{eXtreme Gradient Boosting} (XGB)\footnote{https://xgboost.readthedocs.io/en/latest/} \cite{chen:2016} 
and \textit{CatBoost} (CBT)\footnote{https://catboost.ai} \cite{prokhorenkova:2018}, were implemented. 
Moreover, \textit{Convolutional Neural Network} (CNN) \cite{lecun:2015} was implemented as a deep learning 
algorithm using the Python deep learning API called the Keras Toolkit\footnote{https://keras.io}. Finally, three 
decision-tree based algorithms, \textit{Regularized Greedy Forests} (RGF)\footnote{https://pypi.org/project/rgf-python/} 
\cite{johnson:2014}, \textit{Optimal Tree Ensemble} (OTE) \footnote{https://cran.r-project.org/web/packages/OTE/index.html} 
\cite{khan:2020,khan:21} and \textit{Random Forest} (RF) \cite{breiman:2001}, were applied. \textit{Logistic Regression} (LR) 
was used as the baseline for performance comparison.

\textbf{Sequence-based prediction.} Unlike feature-based prediction, the sequence-based prediction treats the features dependent 
when local context is considered. In this prediction, \textit{Transformer-based Deep Learning} (TDL) \cite{Vaswani:2017} was 
employed for industry sector detection. TDL techniques introduce a novel encoder-decoder architecture in which transformer 
models perform self-attention mechanism to deal with long-range dependencies with ease for the sequence-to-sequence tasks. 
In this study, the BERT (Bidirectional Encoder Representations from Transformers) language model \cite{Devlin:2018} was applied 
because it outperformed other popular transformer models (e.g., \textit{DistilBERT}, \textit{GPT-2}, \textit{XLNet}, and \textit{RoBERTa}) 
in our curated training data.

\section{Evaluations and experimental results}
\label{results}

\subsection{Cross-validation}

To avoid overfitting, stratified 10-fold cross-validation (CV-10) is employed for the building of the predictive model and the 
validation of the models' performance. The inclusion of feature selection in the cross-validation loop avoids potential bias 
as demonstrated in \cite{croner:2005}. 

\begin{figure*}
\includegraphics[scale=0.28]{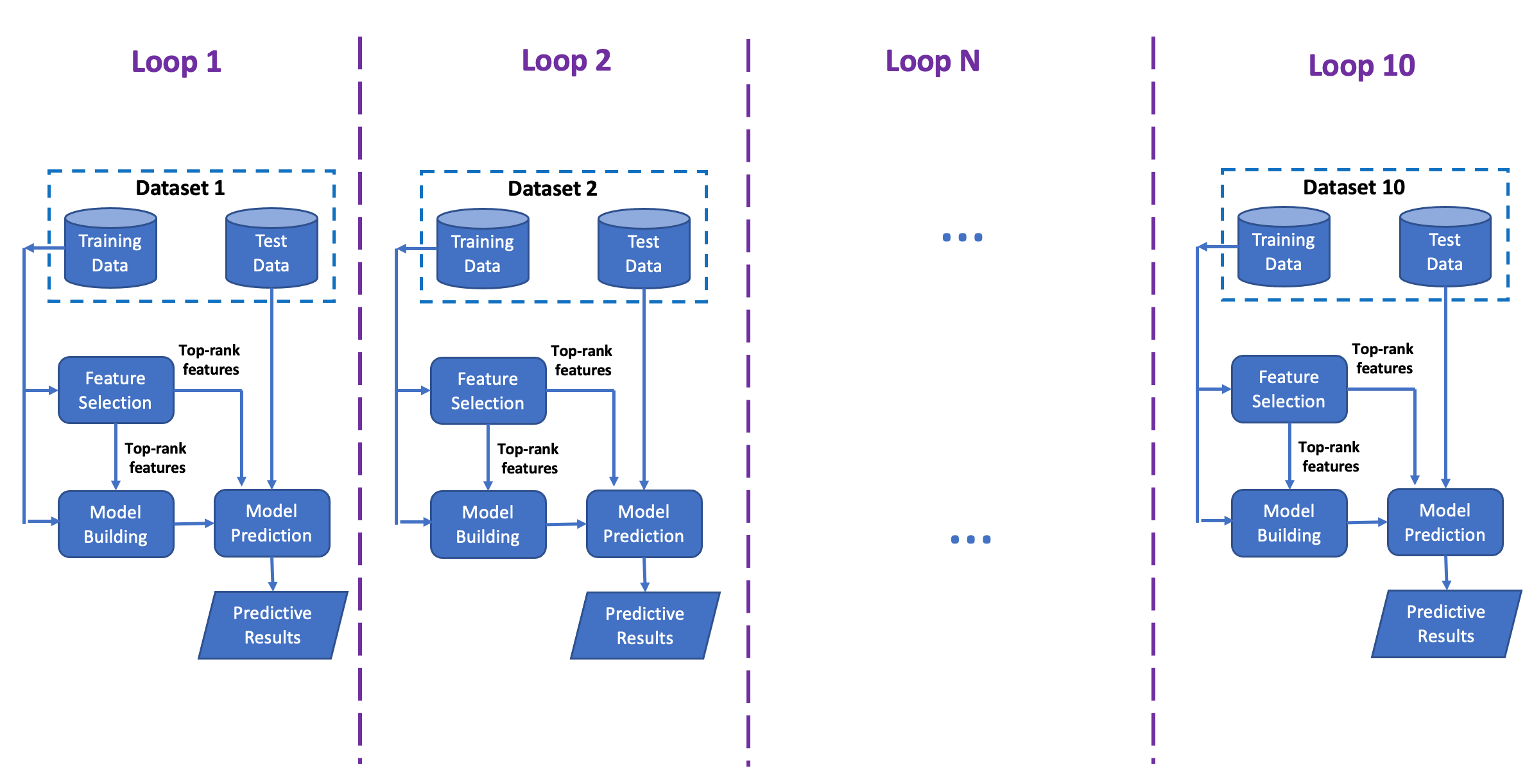}
\caption{Feature selection within the 10-fold cross validation procedure. } 
\label{fig:feature_loops}       
\end{figure*}

As shown in Figure \ref{fig:feature_loops}, for the annotated industry detection dataset, by applying the stratified 
CV-10 method \cite{Airola:2011}, the dataset is randomly split into 10 subsets. Each fold is ensured to have the same 
proportion of instances with the target industry label. At every loop for each unique subset, the subset is the hold out 
test data (10\% of the dataset) and the remaining 9 subsets comprise the training data (90\% of the dataset). Therefore, 
at each iteration of the cross-validation procedure the dataset consists of the training data (90\% data) and test data (10\% data). 
In each loop, the training data is used at two stages: (a) feature selection to obtain top-ranking features; (b) model building to 
get predictive models using different ML algorithms. Then, the predictive models are applied on the test data to output industry 
detection results. For the performance of each ML algorithm on one specific industry sector, the overall performance is the average 
of the AUC scores or F-scores of the predictive results on the test data in terms of the 10 unique subsets. As a result, the 
cross-validation (CV) estimates of $TPR$, $FPR$ and $ROC$ are unbiased \cite{adler:2009}. For a dataset $\mathcal{D}$ 
which comprises of observations $(x_i,y_i)$, the CV estimators of $TPR$, $FPR$ and $ROC$ are given as follows:

\begin{equation}
\widehat{TPR}^{CV}(t) = \frac{1}{n_1} \sum_{(x_i,y_i)\in D_1} I(\hat{f}^{-k_i}(x_i)\geq t), \ \ t \in [0,1], 
\end{equation}

\begin{equation}
\widehat{FPR}^{CV}(t) = \frac{1}{n_0} \sum_{(x_i,y_i)\in D_0} I(\hat{f}^{-k_i}(x_i)\geq t), \ \ t \in [0,1],
\end{equation}
where $D_1, D_0$ are the subsets of instances for which $y_i=1$ and $y_i=0$, respectively and $n_1, n_0$ are the number 
of instances in $D_1$ and $D_0$, respectively. Moreover, 
$I(\cdot)$ denotes the indicator function, $\hat{f}^{-k_i}$ is the ML algorithm trained without the $k_i$th subset which is used 
for testing \cite{adler:2009}. The CV estimation of the ROC curve is then calculated as follows:

\begin{equation}
\widehat{ROC}^{CV}( \cdot ) = { ( \widehat{FPR}^{CV}(t),  \widehat{TPR}^{CV}(t)), \ \ t \in [0, 1]}
\end{equation}

\subsection{Full feature based performance}
\label{subsec:result_fullFeature}
As discussed in Section \ref{subsec:algorithm}, a number of ML algorithms were applied to 
construct predictive models for industry sector detection using either feature-based prediction 
or sequence-based prediction. To examine their prediction performance, a set of experiments 
were conducted using two separate feature sets (see Section \ref{subsec:feature_set}). 

\begin{table*}[t]
\centering
\setlength\tabcolsep{4.5pt} 

\begin{tabular}{|l|llllll||l|}
\hline
	& \multicolumn{7}{c|}{AUC Score} \\
\cline{2-8}
    & Financial & Health & Technology & Property & Energy & Insurance & Mean \\
		&  &  &  &  &  &  & Score \\\hline
XGB & 0.888 & 0.880 & 0.775 & 0.886 & 0.860 & 0.866 & 0.860\\
CBT & 0.893 & 0.901 & 0.833 & 0.909 & 0.894 & 0.877 & 0.895\\
CNN & 0.908 & 0.897 & 0.859 & 0.902 & \textbf{0.917} & 0.882 & 0.912\\
RGF & \textbf{0.915} & \textbf{0.928} & 0.889 & \textbf{0.921} & \textbf{0.912} & 0.910  & \textbf{0.917}\\
RF & 0.900 & \textbf{0.916} & \textbf{0.903} & \textbf{0.923} & \textbf{0.906} & \textbf{0.917} & 0.916\\
OTE & 0.894 & \textbf{0.912} & 0.873 & \textbf{0.929} & \textbf{0.912} & 0.895 & 0.908\\
LR & 0.844 & 0.842 & 0.754 & 0.859 & 0.838 & 0.841 & 0.837\\
TDL & 0.883 & 0.880 & 0.809 & 0.898 & 0.904 & 0.857 & 0.848\\\hline\hline
	& \multicolumn{7}{c|}{F-score} \\
\cline{2-8}
           & Financial & Health & Technology & Property & Energy & Insurance & Mean \\
		&  &  &  &  &  &  & Score \\\hline
XGB & 0.762 & 0.769 & 0.721 & 0.805 & 0.808 & 0.796 & 0.772\\
CBT & 0.785 & 0.788 & 0.762 & 0.815 & 0.821 & 0.778 & 0.794\\
CNN & 0.822 & 0.800 & 0.793 & \textbf{0.827} & \textbf{0.856} & 0.791 & \textbf{0.819}\\
RGF & \textbf{0.838} & 0.803 & 0.795 & 0.826 &  \textbf{0.858} & 0.795 & \textbf{0.818}\\
RF & 0.834 & \textbf{0.809} & \textbf{0.807} & 0.820 &  \textbf{0.853} & \textbf{0.806} & \textbf{0.819}\\
OTE & 0.828 & 0.805 & 0.781 & \textbf{0.831} & 0.842 & 0.782 & 0.815\\
LR & 0.782 & 0.748 & 0.720 & 0.791 & 0.788 & 0.755 & 0.752\\
TDL & 0.810 & 0.764 & 0.768 & 0.786 & 0.737 & 0.702 & 0.729\\\hline
\end{tabular}
\caption{Performance of different ML algorithms, measured in AUC and F scores, using the \emph{full} features set. 
The largest score value, for each industry sector, is shown in bold. Score values which are not statistical significant, 
from the largest value, are also shown in bold; they have been defined by $p$-values of Wilcoxon signed rank 
test $> 0.001$, showing mean values of 1000 re-sampled cross-validations. }
\label{tab:result_fullFeature}
\label{tab:result_fullFeature}       
\end{table*}

For the performance of different ML algorithms in terms of various industry sectors, Table~\ref{tab:result_fullFeature} shows 
the AUC score using the full features set, that is, Feature Set I for the traditional ML methods and Feature Set II for the 
transformer models. The results show that the predictive models performed best in the classification of the industry 
sectors \emph{Property} and \emph{Health}. It also indicated that, using the full features set, the predictive models 
performed well in the classification the industry sector \emph{Energy} which had the best F-score of over 0.85. 
However, the predictive model did not perform as good for detecting the industries \emph{Technology} 
and \emph{Insurance}, with F-scores which are merely above 0.80.

As demonstrated in Table~\ref{tab:result_fullFeature},  the six selected ML algorithms performed differently 
with respect to various industry sectors. 
Overall, given the mean AUC and F-scores, \emph{RGF} and \emph{RF} outperformed the other ML algorithms. 
However, the best-performed algorithm varied depending on the selected target industry.

Additionally, Table~\ref{tab:result_fullFeature} show that the sequence-based prediction TDL models 
did not demonstrate their advantages on this NLP task as expected. Compared with the well-performed 
algorithms, that is RGF and RF, the performance of the TDL models generally dropped down by 
about $3 \sim 10\%$ on the AUC score and $2\sim 5\%$ on F-score in terms of the identification of all 
the six industry sectors. This implies that sequence-based prediction algorithms might not be appropriate 
for the text classification of legal articles at the full-text article level.

\subsection{Top-ranking features for industry detection}
\label{subsec:result_topRankFeature}
\begin{figure}
\includegraphics{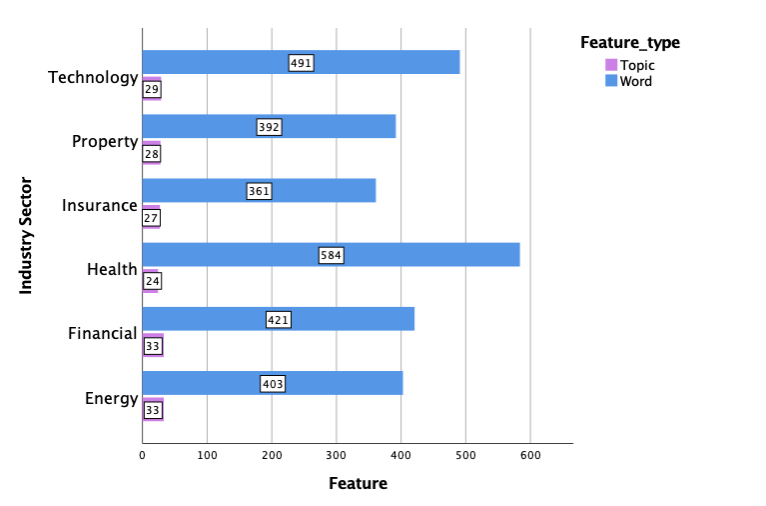}
\caption{The distribution of important features found in individual industries.}
\label{fig:important_feature}       
\end{figure}

It is known that not all the features are important for industry identification. Therefore, it is vital to uncover 
those important features with significant effect on prediction. As described previously, for each industry, a 
feature selection step was applied to generate an integrated list of key features by merging the top-ranking 
features found by different ML algorithms. Figure \ref{fig:important_feature} depicts the number of important 
word and topic features found by the selected algorithms with respect to individual industry sectors.

Additionally, another set of experiments were conducted using the generated important feature lists (see 
Section \ref{subsec:feature_selection}). For one particular industry, the same set of key features was applied 
on the selected ML algorithms to demonstrate the effectiveness of the extracted important features on industry 
prediction compared with the full features (see Table \ref{tab:result_fullFeature}).

\begin{table*}[t]
\centering
\setlength\tabcolsep{4.5pt} 

\begin{tabular}{|l|llllll||l|}
\hline
	& \multicolumn{7}{c|}{AUC Score} \\
\cline{2-8}
           & Financial & Health & Technology & Property & Energy & Insurance & Mean \\
           &  &  &  &  &  &  & Score \\\hline
XGB & 0.886 & 0.858 & 0.779 & 0.890 & 0.860 & 0.884 & 0.855\\
CBT & 0.910 & 0.905 & 0.864 & 0.910 & 0.888 & 0.891 & 0.890\\
CNN & 0.928 & 0.899 & \textbf{0.890} & 0.919 & 0.915 & 0.919 & 0.909\\
RGF & \textbf{0.934} & \textbf{0.913} & 0.889 & 0.930 & \textbf{0.916} & 0.921  & \textbf{0.914}\\
RF & 0.928 & 0.911 & 0.885 & 0.931 & 0.910 & \textbf{0.928} & 0.912\\
OTE & 0.913 & 0.912 & 0.876 & \textbf{0.932} & 0.912 & 0.904 & 0.908\\
LR & 0.861 & 0.837 & 0.739 & 0.876 & 0.847 & 0.863 & 0.834\\
TDL & 0.852 & 0.873 & 0.784 & 0.875 & 0.881 & 0.824 & 0.848\\
\hline\hline
	& \multicolumn{7}{c|}{F-score} \\
\cline{2-8}
           & Financial & Health & Technology & Property & Energy & Insurance & Mean \\
					           &  &  &  &  &  &  & Score \\\hline
XGB & 0.773 & 0.764 & 0.707 & 0.815 & 0.808 & 0.760 & 0.769\\
CBT & 0.794 & 0.782 & 0.769 & 0.817 & 0.821 & 0.778 & 0.792\\
CNN & 0.832 & \textbf{0.819} & 0.787 & 0.827 & 0.856 & 0.791 & \textbf{0.817}\\
RGF & \textbf{0.839} & 0.799 & 0.795 & 0.820 & \textbf{0.858} & 0.795 & 0.816\\
RF & 0.836 & 0.792 & \textbf{0.802} & 0.822 & 0.853 & \textbf{0.806} & 0.816\\
OTE & 0.832 & 0.812 & 0.790 & \textbf{0.828} & 0.842 & 0.782 & 0.814\\
LR & 0.741 & 0.740 & 0.688 & 0.797 & 0.788 & 0.755 & 0.750\\
TDL & 0.748 & 0.734 & 0.729 & 0.725 & 0.737 & 0.702 & 0.729\\
\hline
\end{tabular}
\caption{Performance of different ML algorithms, measured in AUC and F scores, using the selected \emph{important} features. 
For each industry sector, the largest mean score value of 1000 re-sampled cross-validations is shown in bold. }
\label{tab:result_topRankFeature}       
\end{table*}

The AUC and F-score results shown in Table ~\ref{tab:result_topRankFeature} indicates that the selected 
key features achieve competitive, in fact slightly better performance compared with the full feature set. For 
all the six industries, the best prediction performance has an AUC score higher than 0.90 as well as a 
F-score above 0.81. This indicates that the key features extracted and then selected by the algorithms capture 
most of the industry-related characteristics. Also, as in the experimental results with the full features set, the 
models that on average perform the best are the ones which are obtained from the \emph{RGF} and \emph{RF} ML 
algorithms. In terms of the F-score, \emph{CNN} has the highest average score. However, statistical tests were 
conducted to check whether the average F and AUC scores are different between the various algorithms. The results 
showed that the F and AUC scores obtained from CNN are not statistically significant from the \emph{RGF} 
and \emph{RF} algorithms.  

For the sequence-based prediction, given a list of identified important features, for one article, a new text sequence is 
firstly generated by only keeping the key features but filtering out the unimportant ones, which is then fed into the TDL 
models. However, as Table~\ref{tab:result_topRankFeature} show, the TDL models do not work well on the important 
feature based text sequence in terms of both AUC and F-score compared with the performance of the full features and 
also, they under-perform the traditional ML algorithms. The possible explanation for the dropped prediction capability of 
the TDL models is because the newly generated text sequence, by filtering out unimportant features, might lose useful 
contextual information to some extent. The above experiments show that such loss of contextual information plays some 
role in the sequence-based prediction performance.

Overall, in the feature-based prediction, the seven algorithms performed consistently with respect to AUC score and F-score. 
However, the algorithms, \emph{CNN}, \emph{RGF}, \emph{RF} and \emph{OTE}, demonstrated stronger industry prediction 
ability compared to the other selected algorithms. That is, these four algorithms gave slightly better results in terms of the 
performance metrics AUC and F-score, for different industry sectors. 

\section{Discussion}
\label{discussion}

In this work, comparative study was conducted to examine the effectiveness of various ML algorithms on the detection of industry 
sectors on the basis of a small and legal-specific curated dataset. Two types of prediction were employed: feature-based prediction 
with one-dimension weighted features and sequence-based prediction with high-dimension embeded features. The findings and 
lessons learnt from the work are discussed below.

\subsection{Feature selection}
\label{subsec:discuss_featureSelect}

A feature selection step was applied in this study, which made use of the built-in feature importance function of some ML algorithms 
to unearth the industry-specific features of significance.  The results in
Table~\ref{tab:result_topRankFeature} show that the selected important features work well in detecting the various industries. It is 
suggested that a set of important features has the capability of capturing the characteristics of the target industry. Furthermore, 
compared to the full feature set of Section \ref{subsec:result_fullFeature}, the selected important features' set greatly reduced the 
execution time of industry sector prediction by 35\% when working on the computer with a 8-core Intel Core Processor and 16G memory. 

\begin{figure}
\centering
\subfloat[Panel~A: Word cloud for the important \emph{word} features of the \emph{Financial Services} industry.]{
\includegraphics[scale=0.5]{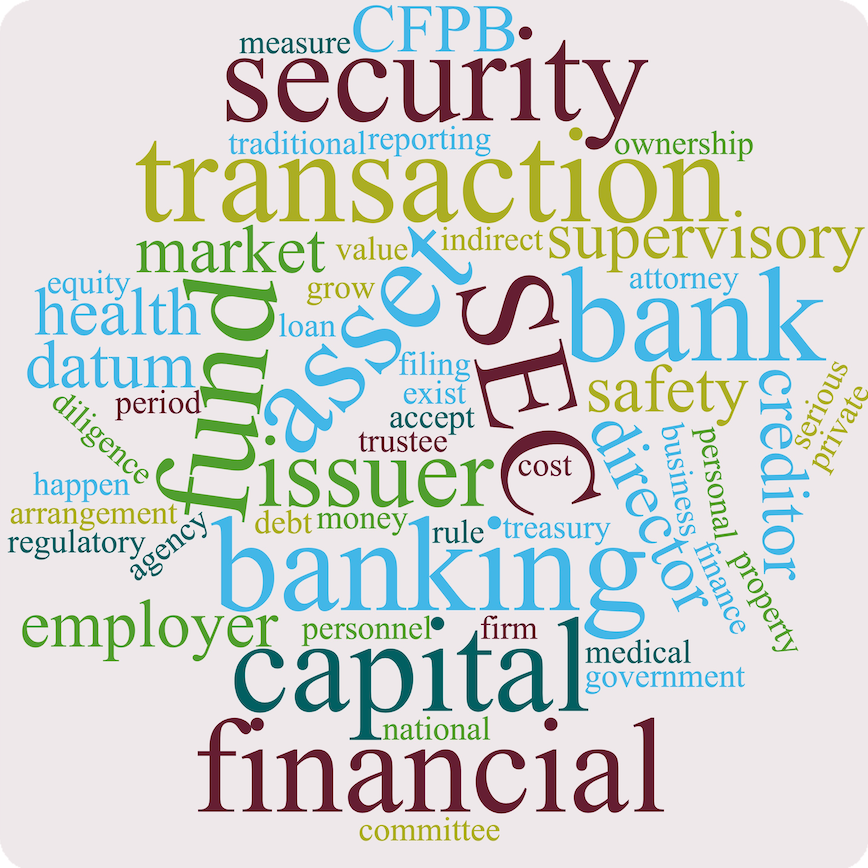} 
\label{fig:wordCloud_financialWord}}
  \hspace{0.5cm}
\subfloat[Panel~B: Word cloud for the important \emph{topic} features of the \emph{Financial Services} industry.]{
\includegraphics[scale=0.5]{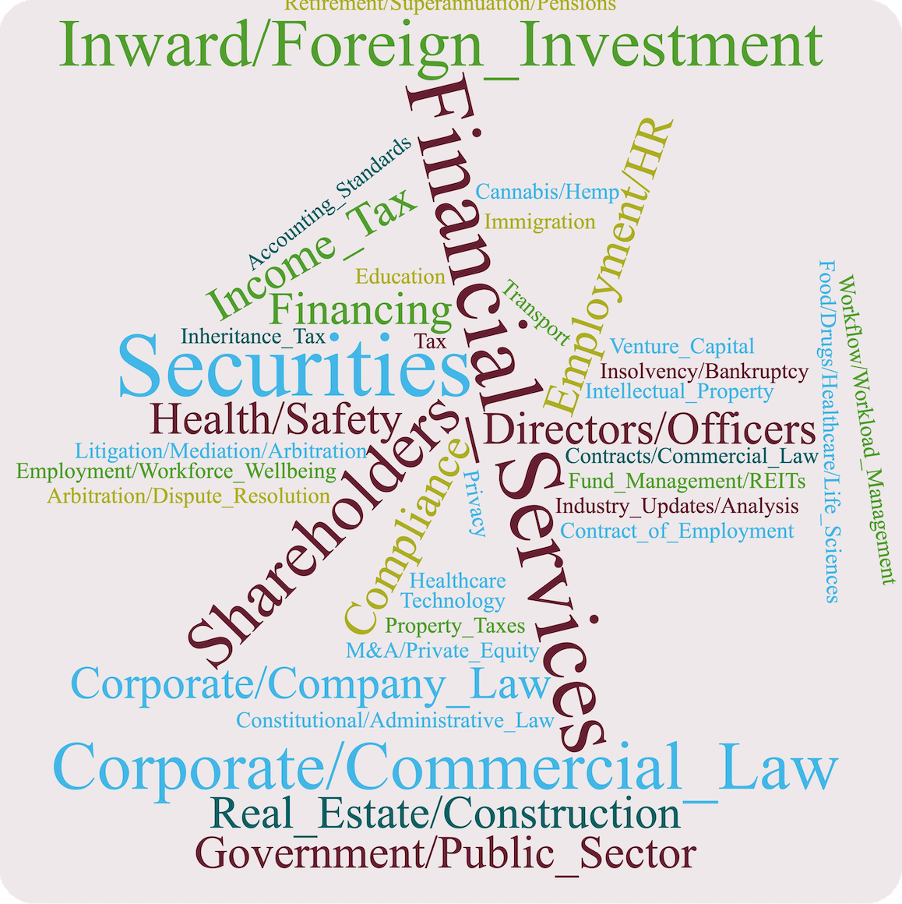} 
\label{fig:wordCloud_financialTopic}}  
\caption{Word clouds of the sample's most important features for the \emph{Financial Services} industry.}
\end{figure}

The number of significant features for individual industries is ranged between 350-600 (see Figure \ref{fig:important_feature}). 
Figures~\ref{fig:wordCloud_financialWord} and \ref{fig:wordCloud_financialTopic} present in a word cloud the sample features 
of importance in the \emph{Financial Services} industry. More information about the important word and topic feature lists in 
individual industry sectors could be found in the supplementary material (\emph{Supplement$\_$material.docx}).

As mentioned in Section \ref{subsec:feature_selection}, each industry had its own ranked feature list with feature importance 
scores. The selected features of importance were generated by integrating the top-ranking features with high score found by 
different ML algorithms. It is important to note that the top-ranking feature lists created by different ML algorithms regarding one 
industry sector were slightly different to some extent. It implied that each ML algorithm might seize different aspects of industry 
characteristics. The results showed that when the important features from different ML algorithms were combined, the feature 
list contained more comprehensive information related to the target industry, and thus the integrated feature list generally 
outperformed the top-ranking list of individual ML algorithms. Moreover, different combinations from several algorithms' 
top-ranking lists were exploited. Overall, there is an overlap, in the range of 66\% to 90\%, of common important features 
between the ML algorithms. The difference in the AUC and F-score performance was in the range of 0.5\%-1.5\%.

\subsection{Industry sectors selected for the study}
\label{subsec:discuss_industrySector}
In this research, six popular industry sectors that frequently occur in the legal articles were selected for the study. 
The F-score performance in Table \ref{tab:result_topRankFeature} shows that the detection of the \emph{Energy} 
industry is the best whereas the prediction on both \emph{Technology} and \emph{Insurance} looks a little worse. 
The possible explanation is that the topics regarding the \emph{Energy} sector are relatively specific and narrow, 
and are better covered by the curated training data. However, the \emph{Technology} sector is connected to a wide 
variety of topics which require more annotated data. In addition, the weak identification of the \emph{Insurance} is 
probably mainly due to the shortage of the annotated data. 
 
Some of the industry sectors, or example \emph{Property} and \emph{Insurance}, are relatively narrow and specific. 
The experience learned by the experiments suggests that one particular industry should have at least 100 full-text articles 
annotated with the corresponding industry tag, sometimes even more especially for the broad industries, to achieve some 
satisfactory performance.

\subsection{\textbf{ML algorithms and techniques used in industry detection}}
\label{subsec:discuss_MLAlgorithm}
The experimental results illustrate that, compared with the sequence-based prediction, the feature-based prediction seems 
more appropriate for the full-text based article classification task. In the feature-based prediction,
deep learning like Convolution Neural Network and decision-tree based techniques have advantages in the correct 
identification of the various industries compared to \emph{Gradient Boosting Machine} (GBM) algorithms (\emph{XGB} 
and \emph{CBT}). However, the algorithms perform inconsistently on various industries, and the best performed algorithm 
depends on the selected target industry. Part of the explanation for the inconsistent performance is due to the different 
characteristics manifested by various industry sectors.

In the feature-based prediction, deep learning technique, \emph{Convolution Neural Network} (CNN), was used for this study. 
In addition, other deep learning techniques, for example Recurrent Neural Network and word embedding, were also tried 
for the model building but they did not work satisfactorily on this dataset. Moreover, various activation functions 
(e.g., \emph{sigmoid}, \emph{tanh}, \emph{linear}, \emph{relu}, and \emph{PReLU}) and optimizers 
(e.g., \emph{SGD}, \emph{RMSprop}, \emph{Adagrad}, and \emph{Adam}) were also tested.

In the sequence-based prediction, transformer-based language models were implemented for the full-text 
based industry detection. Unlike working on the NLP tasks such as multilingual translation and question 
answering, transformer models did not show any advantage in this text classification task as expected. 
There are several possible reasons to explain the unsatisfactory performance of the transformer models:

\begin{itemize}
\item[$\bullet$] At the tokenization stage, a word piece dictionary is used to map the text sequence to a 
feature space with high dimensions. The word piece dictionary is pre-trained on the basis of a large 
general text corporation to capture the contextual semantics between different words. However, the data 
used for this study is a legal-specific dataset in which the semantics between different words in the legal 
contexts might be different from that of general-domain contexts.   

\item[$\bullet$] At the model training stage, the language models such as BERT models are generally large 
and complicated neural networks with hundreds millions of parameters. The parameter weights were 
pre-trained based on some large-scale training datasets from other NLP tasks. As mentioned earlier, 
the annotated training data used for the industry section detection is a relatively small dataset. The 
question is that this dataset might be not big enough to capture the characteristics of the target industry 
sectors at the parameter turning step when the language models were retrained.   

\item[$\bullet$] Transformer models had proven to be state-of-the-art architecture when working on some 
sentence-level NLP tasks such as sentiment analysis \cite{Sayyida:2022} and question answering \cite{Pearce:2021}. 
Sentence-based text sequences facilitate the semantic learning of local contexts for transformer models. Nevertheless, 
the industry section detection task works on the full-text article level. Transformer models might not display powerful 
prediction ability regarding long-distance contexts.  
\end{itemize}

\subsection{\textbf{Computation time and cost}}
\label{subsec:discuss_computation}
The computation time and cost by different ML methods used in this study were also investigated here. Feature 
Set I with one-dimension features used for the feature-based prediction consists of the one-dimension features. 
The feature space of Feature Set I is much smaller than that of Feature Set II with high-dimension features. 
Therefore, the computation time spent for model building and prediction by the ML algorithms used in the 
feature-based prediction is much less than that of the TDL models in the sequence-based prediction when 
using the same computation resource (e.g. CPU and RAM) for model implementation. Given the same 
annotated data, the actual computation time by different ML algorithms at the model training and validation 
stage is provided in the supplementary material (\emph{Supplement$\_$material.docx}). 

For the TDL models used in the sequence-based prediction, due to the high-dimension features (e.g., 
hundreds of dimensions) in Feature Set II and the large pre-trained transformer models like BERT models, 
powerful computation resource (e.g. GPU/TPU and RAM) is required to speed up the building of predictive 
models.

\subsection{Error analysis}
\label{subsec:discuss_errorAnalysis}
Regarding error analysis, there are two main types of error in this classification task, namely, \emph{false 
negatives} which are interpreted as industry-related articles with low probability score, and \emph{false 
positives} which are non-industry articles with relative high probability score.  The \emph{false negatives} 
cases took into account of about 57\% of the prediction errors, and the \emph{false positives} for 43\% 
after manually examining the prediction errors in the six industry sectors.

\emph{False negatives} were generally caused by two possible reasons: (1) Lack of enough positive 
instances for the ML-based models to distinguish industry-specific features, especially for the industries 
which are associated with broad topics, as for example technology. When the industry with one particular 
topic could not be covered by the positive instances, the predictive model lost the capability of recognising 
relevant industry articles. (2) Presence of articles which are short of distinct industry-related features in the 
text, and thus resulting in misclassification. 

\emph{False positives} were possibly caused by mislabeled articles during the annotation stage and by 
some ambiguous articles which didn't contain strong indicators about the relevant industry and caused 
uncertainty to the annotators during the data curation. For such ambiguous articles, the predictive likeliness 
scores by the models  usually fell in the range between 0.4 and 0.6, which is a grey area for the judgment 
of positive industry cases. 

\section{Conclusion and Future Work}
\label{conclusion}
Implicit industry sector information in massive texts provides a novel way to better understand underlying 
semantics within texts, thus facilitating the effective organisation and management of large volumes of text 
data. This study investigated an intelligent approach for automatic industry detection in legal articles using 
Natural Language Processing combined with Machine Learning (ML) techniques. Different types of ML 
algorithms were explored as well as various text and legal features. Two prediction approaches were applied 
for performance comparison of predictive models: feature-based prediction using traditional ML methods with 
one-dimension features, and sequence-based prediction using transformer-based deep learning with high-dimension 
features. The system achieved some encouraging results with AUC scores above 0.90 and F-scores above 0.81 
for the six selected industry sectors. This implies that the machine learning based industry sectors analysis is 
beneficial for the automatic processing of huge collections of text data in a fast and cost-effective way.

Currently, the system identifies six main industry sectors from the legal articles. This work will be extended to the 
detection of more industries by curating more labelled training data. To help minimize the costs of data annotation, 
approaches like Active Learning are to be explored.

Moreover, identifying appropriate industries from a variety of industry sectors, and then assigning correct industry 
tags to the relevant articles is not an easy task even for manual annotation. The annotation from different people 
on the same dataset might differ in their background knowledge and understanding on various topics.  How to 
effectively curate a large-scale training data and how to secure a curated dataset with high quality from different 
annotators merits further investigation.

Future work will consider the use of other industry evidence types, in addition to text-based features (e.g. word 
token) and associated legal topic tags, to further enhance the feature set. Specifically, entity-based evidence 
also has good potential for use in industry detection prediction. This type of entity-based evidence includes 
company names, public bodies, and legal terms. For example, if some company (e.g. \emph{J.P. Morgan}) and 
Government Organisation (e.g. \emph{U.S. Securities and Exchange Commission }- SEC), frequently occur in the 
article, the article is more likely to be connected to the \emph{Financial Services} industry sector.  Identifying these 
entity names from the text and collecting relevant industry background information associated with the entities will 
be investigated in future work. Such evidence will help enrich the existing feature set for the building of the predictive 
models. 

\bmhead{Acknowledgments}
This work was supported by Innovate UK under the Knowledge Transfer Partnership program with Partnership 
No. KTP011976 ) and Mondaq LTD (a company registered in England, number 02906568). 

\section*{Declarations}
\textbf{Conflicts of interest.} The authors declare that they have no conflict of interest.


\bibliography{IndustryDetection_bibliography_abbrievate}


\end{document}